\documentclass[10pt,twocolumn,letterpaper]{article}

\usepackage[pagenumbers]{cvpr} 

\usepackage{graphicx}
\usepackage{amsmath}
\usepackage{ulem}
\usepackage{amssymb}
\usepackage{booktabs}
\usepackage{subcaption}
\usepackage{threeparttable}
\usepackage{makecell}
\usepackage{multirow}
\usepackage{float}
\usepackage{bbding}
\usepackage{enumitem}
\usepackage[dvipsnames,table]{xcolor} 

\usepackage[pagebackref,breaklinks,colorlinks]{hyperref}
\usepackage[accsupp]{axessibility}  

\usepackage[capitalize]{cleveref}
\crefname{section}{Sec.}{Secs.}
\Crefname{section}{Section}{Sections}
\Crefname{table}{Table}{Tables}
\crefname{table}{Tab.}{Tabs.}

\usepackage{xspace} 

\makeatletter
\DeclareRobustCommand\onedot{\futurelet\@let@token\@onedot}
\def\@onedot{\ifx\@let@token.\else.\null\fi\xspace}

\def\ie{\emph{i.e}\onedot}

\makeatother

\usepackage{pifont}

\def\cvprPaperID{*} 
\def\confName{CVPR}
\def\confYear{2023}

\begin{document}

\title{FreeSeg: Unified, Universal and Open-Vocabulary Image Segmentation }

\author{Jie Qin$^{1,2,3}$$^{\star}$ \quad 
Jie Wu$^{2 \star}$ \quad
Pengxiang Yan$^{2}$ \quad
Ming Li$^{2}$ \quad
Ren Yuxi$^{2}$ \quad
Xuefeng Xiao$^{2}$ \\
Yitong Wang$^{2}$ \quad
Rui Wang$^{2}$ \quad
Shilei Wen$^{2}$ \quad
Xin Pan$^{2}$ \quad
Xingang Wang$^{1 \dagger}$ \\
$^{1}$Institute of Automation, Chinese Academy of Sciences \ 
$^{2}$ByteDance Inc \\
$^{3}$School of Artificial Intelligence, University of Chinese Academy of Sciences
}

\twocolumn[{%
\renewcommand\twocolumn[1][]{#1}%
\maketitle
\begin{center}
    \centering
    \vspace{-20pt}
    \includegraphics[width=\textwidth]{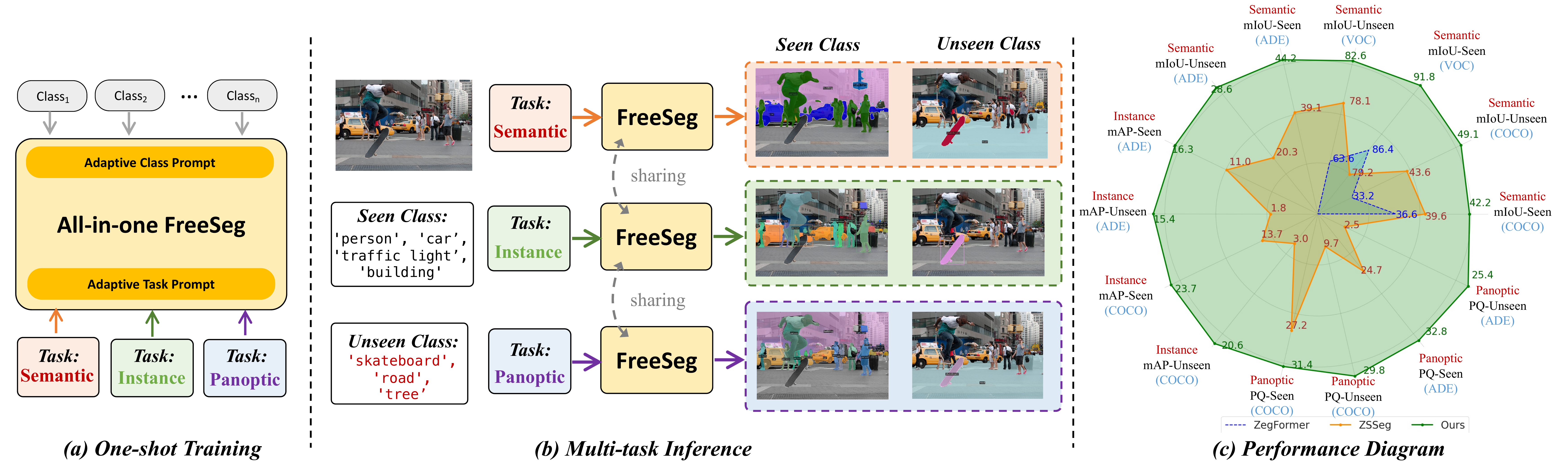}
    \vspace{-10pt}
\captionof{figure}{We propose FreeSeg, a generic framework to accomplish Unified, Universal and Open-Vocabulary Image Segmentation. (a) FreeSeg optimizes an all-in-one network via one-shot training. (b) FreeSeg employs the same architecture and parameters to handle diverse segmentation tasks seamlessly in the inference procedure. (c) FreeSeg establishes new state-of-the-art performance across \textcolor{Bittersweet}{diverse segmentation tasks}, \textcolor{Periwinkle}{training datasets} and zero-shot generalization. 
}
    \label{fig:head}
\end{center}%
}]

\maketitle
\renewcommand{\thefootnote}{}
\footnotetext{$^\star$Equal contribution. $^\dagger$Corresponding author.
This work was done while Jie Qin interned at ByteDance. 
}

\begin{abstract}
Recently, open-vocabulary learning has emerged to accomplish segmentation for arbitrary categories of text-based descriptions,  which popularizes the segmentation system to more general-purpose application scenarios.
However, existing methods devote to designing specialized architectures or parameters for specific segmentation tasks. These customized design paradigms lead to fragmentation between various segmentation tasks, thus hindering the uniformity of segmentation models.
Hence in this paper, we propose \textbf{FreeSeg}, a generic framework to accomplish \textbf{Unified}, \textbf{Universal} and \textbf{Open-Vocabulary} Image Segmentation.
FreeSeg optimizes an all-in-one network via one-shot training and employs the same architecture and parameters to handle diverse segmentation tasks seamlessly in the inference procedure.
Additionally, adaptive prompt learning facilitates the unified model to capture task-aware and category-sensitive concepts, improving model robustness in multi-task and varied scenarios.
Extensive experimental results demonstrate that FreeSeg establishes new state-of-the-art results in performance and generalization on three segmentation tasks, which outperforms the best task-specific architectures by a large margin: \textbf{5.5\%} mIoU on semantic segmentation, \textbf{17.6\%} mAP on instance segmentation, \textbf{20.1\%} PQ on panoptic segmentation for the unseen class on COCO.  Project page: \href{https://FreeSeg.github.io}{https://FreeSeg.github.io}.

\end{abstract} 

\section{Introduction}
\label{sec:intro}

Image segmentation has been one of the most widely researched topics in computer vision, aiming to simultaneously group and categorize object pixels in the image.
In the recent literature, the image segmentation community has witnessed tremendous success at cost of large-scale datasets\cite{chen2017deeplab, sun2019deep,bucher2019zero}, where objects are exhaustively annotated with pixel-level masks and category labels.
However, due to the time-consuming and laborious annotations, the template categories sizes of existing segmentation tasks are still limited to an order of $10$ or $10^2$,  which is in orders of magnitude much smaller than the vocabulary that humans use to describe the real world.
Such learning objective binds the segmentors’ scalability into a limited cognitive space, and it becomes a critical bottleneck when this system is popularized to handle richer and more generalized semantics.

As a viable path to handle categories of custom specification beyond the training dataset, open-vocabulary learning leverages large-scale visual-language pre-training models (such as CLIP\cite{radford2021learning}, ALIGN\cite{jia2021scaling}) to calculate matching similarity between visual concept and text corpus.
Recently, a series of segmentation-based open-vocabulary studies\cite{xian2019semantic, bucher2019zero, xu2021simple} have emerged to design task-specific architectures and parameters for individual segmentation task.
For example, ZSSeg\cite{xu2021simple} leverages the off-the-shelf pre-trained CLIP model and achieves competitive performance in open vocabulary semantic segmentation. However, current works suffer from two obvious shortcomings when popularized to general segmentation scenes: 
i) \textit{task-insensitive}: they can not capture task-aware characteristics and be effectively generalized to diverse segmentation tasks;
ii)  \textit{resource-unfriendly}: the model needs to be trained from scratch when switching tasks, and diverse tasks require deploying multiple customized models.
Although MaskFormer\cite{cheng2021per} succeeds in accomplishing multiple segmentation tasks into one compact system, it still needs to train a customized model for each task and it is not designed for open-vocabulary tasks.
These observations motivate us to raise a question:  
\textit{how to design a unified open-vocabulary framework to accomplish universal segmentation tasks?}

To address the above question, As shown in Fig.\ref{fig:head}, we propose \textbf{FreeSeg}, a novel framework to accomplish \textbf{Unified}, \textbf{Universal} and \textbf{Open-Vocabulary} Image Segmentation. In FreeSeg, our goals are mainly three-fold: i)  Unified: FreeSeg designs a unified (all-in-one) network that employs the same architecture and inference parameters to handle multiple segmentation tasks; ii) Universal: FreeSeg adapts to various tasks, namely semantic, instance and panoptic segmentation; iii) Open-Vocabulary: FreeSeg is capable of generalizing to arbitrary segmentation categories.

In general, FreeSeg advocates a two-stage segmentation framework, with the first stage extracting universal mask proposals and the second stage accomplishing zero-shot classification on these masks.
Specifically, FreeSeg conducts a one-shot training procedure to optimize a unified segmentation model with multi-task labels, which helps to capture task-special characteristics for universal segmentation.
An adaptive prompt learning scheme is introduced to encode task-aware and category-sensitive concepts into the text abstraction.
It enables FreeSeg to flexibly accomplish different segmentation tasks of arbitrary categories, handling all tasks and categories in one model.
To sum up, FreeSeg is a \textit{task-flexible, category-arbitrary and performance-excellent} framework, the main contributions of our work are listed as follows:

\begin{itemize} [ itemsep = 1.0pt]

\item  To the best of our knowledge, we offer the first attempt to tackle a novel computer vision task, namely, unified open-vocabulary segmentation. A universal framework FreeSeg is proposed to employ an all-in-one model with the same architecture and inference parameters to accomplish open-vocabulary semantic, instance, and panoptic segmentation.

\item  Adaptive prompt learning explicitly encodes
multi-granularity concepts (task, category) into compact textual abstraction and helps the unified model generalize to arbitrary text descriptions.
FreeSeg further designs the semantic context interaction and test time prompt tuning mechanism to improve cross-model alignment and generalization for unseen classes.

\item We evaluate FreeSeg on three image segmentation tasks (semantic, instance, and panoptic segmentation) using COCO, ADE20K and VOC 2012. As shown in Fig.\ref{fig:head} (c), extensive experiments demonstrate that FreeSeg establishes new state-of-the-art results in terms of performance and generalization.
In addition to reducing the research effort by at least three times, it outperforms the best-specialized architectures and is more feasible for multi-task deployment.

  \end{itemize}


\section{Related Work}

\subsection{Open Vocabulary Segmentation}
Deep learning ~\cite{li2020pams, ren2021online, xia2022trt, li2019oicsr, wu2020tree, xiao2017design, wu2020fine} and image segmentation has recently witnessed tremendous success~\cite{chen2017deeplab, chen2018encoder, yang2018denseaspp, sun2019deep, cheng2021per, qin2022multi, qin2022activation}.
Open vocabulary segmentation aims to segment the target categories that can not access during the training procedure.
The existing approaches can be divided into two aspects: mapping visual features into semantic space \cite{xian2019semantic, bucher2019zero, gu2020context} and cross-modal alignment with pre-trained models \cite{li2022language, ding2022decoupling, xu2021simple}. 
For the mapping aspect, SPNet\cite{xian2019semantic} encodes visual features to the semantic embedding space and then projects each pixel feature to predict probabilistic outcomes through a fixed semantic word encoding matrix.
ZS3Net\cite{bucher2019zero}  generates the pixel-level features of unseen classes in the semantic embedding space and adopts the generated features to supervise a visual segmentation model. 
STRICT\cite{pastore2021closer} introduces a self-training technique into SPNet to improve the segmentation performance of unseen classes. 
Cross-modal alignment employs robust zero-shot capabilities of the pre-trained cross-modal models such as CLIP\cite{radford2021learning}
to conduct open vocabulary segmentation tasks. 
LSeg\cite{li2022language} learns a CNN model to compute per-pixel image features to match with the text embeddings embedded by the pre-trained text model. 
ZegFormer\cite{ding2022decoupling} and ZSSeg\cite{xu2021simple} leverage the visual model to generate the class-agnostic masks, and use the pre-trained text encoder to retrieve the unseen class masks. 
XPM\cite{huynh2022open} utilizes the region-level features to match CLIP-based text embeddings to accomplish the open vocabulary instance segmentation. MaskCLIP\cite{ding2022open} attempts to establish relationships between the class-agnostic masks in the CLIP visual encoder to complete the open vocabulary panoptic segmentation.

\subsection{Universal Segmentation Architecture}
The goal of the universal segmentation framework is to employ the same architecture in arbitrary segmentation tasks, so current universal segmentation approaches\cite{cheng2021per, zhang2021k, cheng2022masked} regularly constrain multiple tasks (\textit{semantic}, \textit{instance}, \textit{panoptic}) to a unified training paradigm.
MaskFormer\cite{cheng2021per} unifies the segmentation tasks into a classification problem for masks, \ie, outputting binary masks and the corresponding categories, which achieves state-of-the-art performance in both semantic and panoptic segmentation tasks. 
K-Net\cite{zhang2021k} standardizes instance segmentation into semantic segmentation via learnable kernels to accomplish the semantic, instance, and panoptic segmentation tasks simultaneously. Mask2Former\cite{cheng2022masked} employs the masked attention mechanism into MaskFormer to improve the generalization of the unified model and the performance of each task. However, these unified frameworks still require training a separate model for each task to achieve the best performance. Our proposed FreeSeg conduct one-shot training to optimize an all-in-one model to finish multiple segmentation tasks.

\subsection{Prompt Learning}
Prompt learning achieved a remarkable leap in the field of NLP \cite{hu2021lora, lester2021power, wei2021finetuned}, and then is rapidly popularized into the vision or vision-language models\cite{zhou2022learning, rao2022denseclip}. CoOp\cite{zhou2022learning} brings continuous prompt optimization from downstream data to adapt the pre-trained vision-language model. 
DenseCLIP\cite{rao2022denseclip} finetunes the pre-trained text encoder with the given prompt templates to perform text and visual feature matching for downstream intensive prediction tasks such as detection and segmentation. 
For open vocabulary segmentation tasks\cite{ding2022decoupling, xu2021simple}, prompt templates are generated from the given category names, and then are encoded to the text embeddings for matching the unseen classes.

\section{Methodology}

\subsection{FreeSeg Framework}

\begin{figure*}[t]
    \centering
    \begin{center}
        \includegraphics[width=1.0\textwidth]{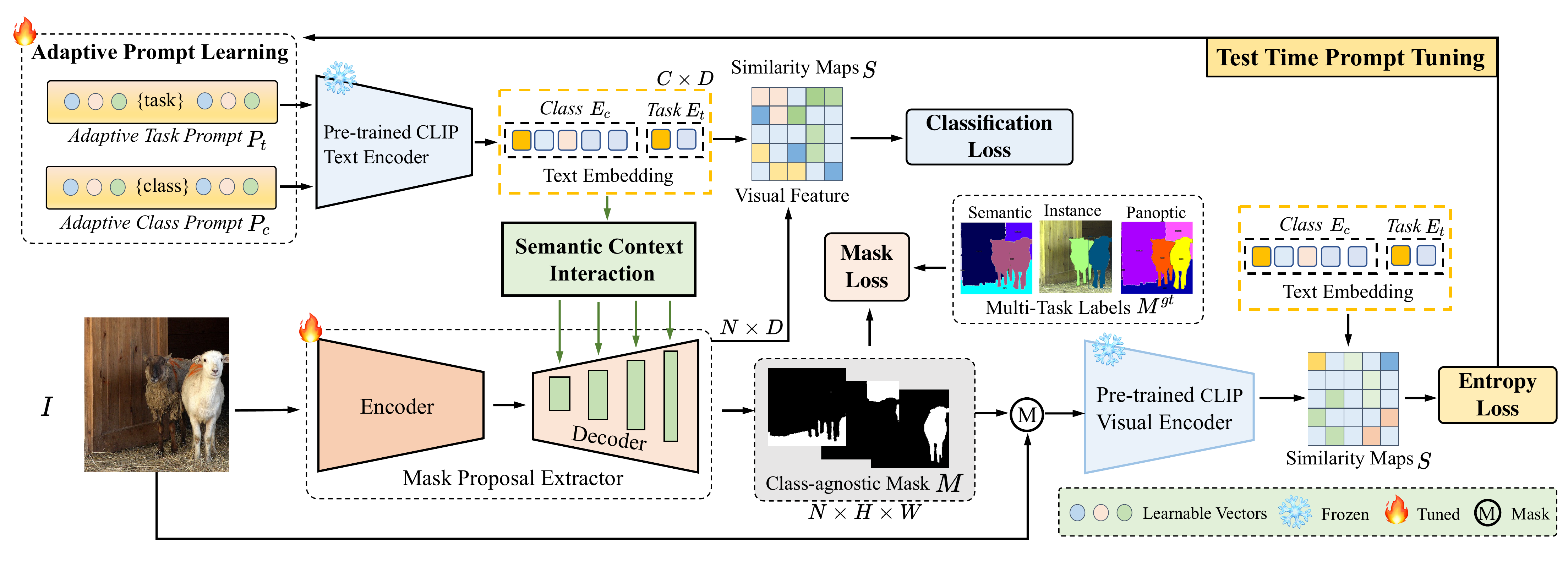}
    \end{center}
        \vspace{-20pt}
    \caption{ Overview of our two-stage FreeSeg framework. 
i) one-shot training: optimizes an all-in-one segmentation model via multi-task supervision to generate universal mask proposals; ii) Multi-task inference: leverages pre-trained CLIP to classify mask proposals according to adaptive task and class prompt.}
    \label{fig: framework}
    \vspace{-5pt}
    \end{figure*}

The proposed unified open-vocabulary segmentation aims to optimize an all-in-one model to obtain semantic, instance, and panoptic segmentation results on arbitrary categories.
To address this novel task,  we propose a novel framework to accomplish unified and universal open vocabulary segmentation in this paper, termed as FreeSeg.
FreeSeg advocates a two-stage framework, with the first stage extracting universe mask proposals and the second stage leveraging CLIP to perform zero-shot classification on the masks which are generated in the first stage. The whole framework of FreeSeg is illustrated in Fig. \ref{fig: framework}.

\textbf{Training.} The training data in the first stage
contains images $I$, seen category set $C_{seen}$, task names $T_{train}$ and multi-task labels $M^{gt}$. 
The training procedure only accesses the seen categories $C_{seen}$ and the corresponding labels.
The mask proposal extractor encodes the image into visual concepts $F_v \in \mathcal{R}^{N \times D}$ and class-agnostic masks $M \in \mathcal{R}^{N \times H \times W}$, where $N$ and $D$ denote the number of queries and feature dimensions. 
To encapsulate multiple learned tasks in a unified model, We leverage three task-specific labels, \ie, $M^{gt}\in (M^{gt}_{sem}, M^{gt}_{ins}, M^{gt}_{pan})$ to selectively supervise the mask proposal extractor with mask loss:
\begin{equation}
 \mathcal{L}_{mask} = \mathcal{L}_F(M, M^{gt}) + \mathcal{L}_D(M, M^{gt})
\text{,} 
    \end{equation}
where $\mathcal{L}_F$ denotes the Focal\cite{lin2017focal} loss and $\mathcal{L}_D$ is the Dice\cite{milletari2016v} loss. 
Simultaneously optimizing all tasks is often difficult due to gradient conflicts across tasks during training, thus only one task label is selected for supervision per iteration, which is randomly selected from the $(M^{gt}_{sem}, M^{gt}_{ins}, M^{gt}_{pan})$.

To facilitate FreeSeg to handle task and categories characteristics, we design a novel adaptive prompt learning to explicitly embed task and category concepts into joint text embeddings $F_t \in \mathcal{R}^{C \times D}$ via a pre-trained CLIP-based text encoder, where $C$ denotes the number of categories. 
The cross-modal classification supervision is set up to enable FreeSeg to classify generated masks according to arbitrary text.
Specifically, the visual concepts $F_v$ are leveraged to compute the similarity matching map with text embeddings $F_t$. The cosine similarity score $\mathcal{S} \in \mathcal{R}^{N \times C}$ between pairs of $F_v^i$ and $F_t^j$  is computed as:
\begin{equation}
    S(i, j) = cos(F_v^i, F_t^j) =\dfrac {F_v^i \cdot F_t^j} {\left\| F_v^i\right\| \left\| F_t^j\right\|} 
\text{,} 
    \end{equation}
where $i\in[1, N]$, $j\in[1,C]$.
The obtained similarity matching map indicates the probability of the predicted category for all class-agnostic masks, which is supervised by the class labels with the cross-entropy loss $\mathcal{L}_{cla}$.
The total training loss is formulated as:
\begin{equation}
    \mathcal{L} = \mathcal{L}_{cla} + \mathcal{L}_{mask}
\text{,} 
    \end{equation}

\textbf{Testing.}
In the testing phase, the trained mask proposal extractor generates a set of binary masks with textual guidance and leverages the pre-trained CLIP visual encoder to obtain mask-level visual concepts.
FreeSeg calculates the similarity between mask representation and compact text embedding and outputs task-oriented segmentation results according to the adaptive task prompt.
With the aid of adaptive prompt learning, FreeSeg can handle arbitrary tasks and categories. The test category set $C_{test}$ consists of seen classes $C_{seen}$ and additional unseen classes $C_{unseen}$.

\subsection{Adaptive Prompt Learning}
To encode arbitrary tasks and categories into compact textual abstraction, we propose the adaptive prompt learning module containing the adaptive task prompt $P_t$ and the adaptive class prompt $P_c$. 
Fixed prompt puts all category and task names into the same templates, which is not the optimal representation for task-category pair contexts. 
While adaptive prompt learning turns the task and category texts into a set of learnable vectors, which are concatenated as text embeddings to facilitate model training.

\textbf{Adaptive Task Prompt}. 
The adaptive task prompt promotes capturing task-specific characteristics, encapsulating multiple learned tasks in a unified framework, and effectively disentangles the parameter spaces to avoid different tasks’ training conflicts.
Specifically, the adaptive task prompt $P_t$ is generated according to the template \textit{\{$\circ$ $\circ$ ... $t$ ... $\circ$ $\circ$ \} }, where $\circ$ denotes the learnable vectors.
$t$ is the corresponding task name in a task set $T$, which contains ``\textit{semantic segmentation.}'', ``\textit{instance segmentation.}'', or ``\textit{panoptic segmentation.}''. Then the task prompts are embedded by the pre-trained CLIP text encoder $\varPsi$:
\begin{equation}
    E_{t} = \varPsi (P_t(t)), t \in T \text{,}
    \end{equation}
where $E_{t}$ denotes the task embeddings.

\textbf{Adaptive Class Prompt}. 
An adaptive class prompt is introduced to popularize FreeSeg to generalize to broader unseen categories and improve open-domain performance.
Given the semantic categories $C_{seen}$ involved in training, the class prompts $P_c$ are obtained by the template \textit{\{$\circ$ $\circ$ ... $c$ ... $\circ$ $\circ$ \} }, where $c$ is the filled class names. 
The adaptive class prompt $P_c$ is embedded to generate the class text embeddings $E_c$:
\begin{equation}
    E_{c} = \varPsi  (P_c(c)), c \in C_{seen} \text{,}
    \end{equation}

To model a joint task-category textual space, the class text embeddings $E_c$ and the task text embeddings $E_t$ are fused to get the multi-granularity embeddings $F_t$:
\begin{equation}
    F_t = Cat(E_{c}, E_{t})   \text{,}
    \end{equation}
where $Cat$ denotes the concatenation operation. 
It is worth noting that the input category can be arbitrary, so $F_t$ can seamlessly adapt to unseen categories for open vocabulary segmentation.

\subsection{Semantic Context Interaction}
\label{sec: interaction}
The vanilla visual concepts ignore task and category information that can provide more reliable cues for comprehensive inference. To address this issue, we creatively introduce a \textit{semantic context interaction module} to improve the cross-modal feature matching and alignment by effectively aggregating adaptive textual embedding into visual concepts.
Specifically, the semantic context interaction module employs the cross-attention module to model the correlations between text embeddings and multiple-scale visual features.
\begin{equation}
  Attn(Q^z, K, V) = softmax(\frac{Q^zK^T}{\sqrt{d_k}})V^T   \text{,}
  \end{equation}
\begin{equation}
    Q^z = \phi_q (F_{v}^z)\text{,} \quad K = \phi_k (F_t)\text{,}  \quad V = \phi_v (F_t) \text{,}
    \end{equation}
where $F_{v}^z$ denotes $z$-layer visual feature from decoder in mask proposal extractor.
$Q^z, K, V$ denote the query, key, and value embeddings generated by the projection layers $\phi_q, \phi_k, \phi_v$. $\sqrt{d_k}$ represents the scaling factor. Then the attention relationship is utilized to enhance the visual features:
\begin{equation}
  \hat{F_{v}^z} = \mathcal{H} \{Attn[\phi_q (F_{v}^z),\phi_k (F_t), \phi_v (F_t)]\} \text{,}
  \end{equation}
where $\mathcal{H}$ denotes the output projection layer. The enhanced visual feature $\hat{F_{v}^z}$ is beneficial to emphasize the visual feature concerning the given text classes.

\begin{table*}[tp]
  \centering
  \footnotesize
  \vspace{-10pt}
  \caption{Comparison with state-of-the-art methods in open vocabulary semantic segmentation. mIoU$^s$ and mIoU$^u$ denote the mIoU(\%) of seen classes and unseen classes. The variant ‘‘Full Sup." denotes training FreeSeg with all seen and unseen classes.  }
   \vspace{-10pt}
  \begin{threeparttable}
  \resizebox{0.9\textwidth}{!}{
    \begin{tabular}{l|ccc|ccc|ccc}
      \toprule
      \multirow{2}{*}{\textbf{Method}} &\multicolumn{3}{c|}{COCO} & \multicolumn{3}{c|}{VOC2012} & \multicolumn{3}{c}{ADE20K}\\ 
      & \textbf{mIoU$^s$} & \textbf{mIoU$^u$} & \textbf{hIoU} & \textbf{mIoU$^s$} & \textbf{mIoU$^u$} & \textbf{hIoU} & \textbf{mIoU$^s$} & \textbf{mIoU$^u$} & \textbf{hIoU} \\ 
      \midrule
      Full Sup. & 42.9 & 54.3 & 47.9 & 92.3 & 89.5 & 91.1 & 46.1 & 41.5 & 44.0 \\ 
      \midrule
      SPNet\cite{xian2019semantic} & 34.6 & 26.9 & 30.3 & 77.8 & 25.8 & 38.8 & - & - & -\\
      ZS5\cite{bucher2019zero} & 34.9 & 10.6 & 16.2 & 78.0 & 21.2 & 33.3 & - & - & - \\
      CaGNet\cite{gu2020context} & 35.6 & 13.4 & 19.5 & 78.6 & 30.3 & 43.7 & - & - & -\\
      STRICT\cite{pastore2021closer} & 35.3 & 30.3 & 32.6 & 82.7 & 35.6 & 73.3 & - & - & -\\
      ZegFormer\cite{ding2022decoupling} & 36.6 & 33.2 & 34.8 & 86.4 & 63.6 & 73.3 & - & - & -\\
      ZSSeg\cite{xu2021simple} & 39.6 & 43.6 & 41.5 & 79.2 & 78.1 & 79.3 & 39.1 & 20.3 & 31.6 \\
      \rowcolor{gray!10} Ours & \textbf{42.2} & \textbf{49.1} & \textbf{45.3} & \textbf{91.8} & \textbf{82.6} & \textbf{86.9} & \textbf{44.2} & \textbf{28.6} & \textbf{39.8} \\
      \bottomrule
      \end{tabular}}
      \end{threeparttable}
      \label{tab:semseg}
  \end{table*}

\subsection{Test Time Prompt Tuning}
To improve the  cross-modal alignment of unseen categories, we leverage the test time adaptation (TTA) algorithm \cite{kundu2020universal, wang2020fully, sun2020test} to refine the adaptive class prompt during testing, termed as \textit{Test Time Prompt Tuning}.
 
In the testing phase, we filter out the cosine similarity scores $S_u$ of unseen classes and calculate the corresponding entropy:
\begin{equation}
  entro = -\frac{1}{N_{u}} \sum^{N_{u}}_{i=1} s_i log(s_i) \text{,}
  \end{equation}
where $entro$ denotes the entropy value of each sample. $N_{u}$ is the number of the unseen classes and $s_i$ is the score of $i^{th}$ class of $S_u$. Then we select the high-confidence queries according to the entropy $S^*_u = S_u[entro < \tau]$, where $\tau$ is the threshold of the high confidence. Because the low entropy value indicates the high confidence level of the sample predictions. We calculate the entropy loss $\mathcal{L}_{ent}$ to optimize the parameters of the adaptive class prompt:
\begin{equation}
  \mathcal{L}_{ent}  = -\frac{1}{N_{u}K} \sum^{N_{u}}_{i=1} \sum^{K}_{j=1} s_{ij} log(s_{ij}) \text{,}
  \end{equation}
where $s_{ij}$ denotes the score of $j$-th  selected queries. $K$ is the queries number in $S^*_u$.

\section{Experimental Results}

\subsection{Datasets and Evaluation Metrics}

\subsubsection{Datasets}

\textbf{COCO.} COCO dataset \cite{lin2014microsoft} contains multi-tasks ground-truth labels towards the same image. We collect semantic labels of COCO stuff\cite{caesar2018coco} and panoptic labels of COCO and merge them to get the unified, category-wide annotations $M^{gt}$. 
We follow \cite{xian2019semantic, xu2021simple} to divide all 171 categories into 156 seen and 15 unseen classes to complete the open vocabulary segmentation task. 

\noindent \textbf{ADE20K.} ADE20K \cite{zhou2017scene} contains 20,000 training images and 2,000 validation images with 150 categories. We split 15 categories into unseen classes, and the remaining 135 are treated as seen/training classes.

\noindent \textbf{PASCAL VOC2012.} We conduct experiments on PASCAL VOC2012\cite{everingham2012pascal} to accomplish semantic segmentation.
Following \cite{xian2019semantic, xu2021simple}, we divide 20 foreground classes into 15 seen classes and 5 unseen classes to evaluate the effectiveness of the open vocabulary segmentation.

\subsubsection{Evaluation Metrics}
\textbf{Semantic segmentation.}
We follow \cite{ding2022decoupling, xu2021simple} to adopt the mean of Interaction over Union (mIoU) to respectively evaluate the open vocabulary semantic segmentation performance for seen and unseen classes. We also employ the harmonic mean IoU (hIoU) among the seen and unseen classes to measure comprehensive performance. 

\noindent \textbf{Instance segmentation.}
We report the mean Average Prediction (mAP) of seen and unseen classes for open vocabulary instance segmentation.

\noindent \textbf{Panoptic segmentation.}
For open vocabulary panoptic segmentation, we follow the setting of fully supervised panoptic segmentation and use task-aware metrics (\ie, PQ, SQ, RQ) to evaluate panoptic segmentation quality.

\subsection{Implementation Details}

\noindent \textbf{COCO.}
We employ Mask2Former\cite{cheng2022masked} as the mask proposal extractor and ResNet101 as the backbone. VIT-B/16 is adopted as the backbone of CLIP \cite{radford2021learning}.  All experiments are conducted on 8$\times$A100 GPUs. 
We take the batch size of 32 per GPU and set the input image size as 640$\times$640. 
The optimizer is AdamW with a learning rate of 0.0002 and weight decay of 0.0002. The number of training iterations is 60,000. 
In addition, the learnable parameter size of the task prompt is 8$\times$512, and the class prompt is 16$\times$512. 
We follow the comparison methods~\cite{pastore2021closer,ding2022decoupling,xu2021simple} to employ the self-training technique for training.

\noindent \textbf{ADE20K and PASCAL VOC2012.}
ADE20K dataset uses  512$\times$512 input image size and the number of iterations is set to 20,000 on PASCAL VOC2012. The remaining training settings on these two datasets are the same as COCO.

\subsection{Comparison to State-of-the-art Methods}

\paragraph{Open Vocabulary Semantic Segmentation}
We compare FreeSeg with current state-of-the-art open vocabulary semantic segmentation methods in Tab.\ref{tab:semseg},  including SPNet\cite{xian2019semantic}, ZS5\cite{bucher2019zero}, CaGNet\cite{gu2020context}, STRICT\cite{pastore2021closer}, ZegFormer\cite{ding2022decoupling}, ZSSeg\cite{xu2021simple}.
Tab.\ref{tab:semseg} can be summarized as the following observations:
i) FreeSeg achieves 49.1\% and 28.6\% mIoU towards unseen classes on COCO and ADE20K, which surpasses the previous best method ZSSeg by +5.5\%  and +8.3\%, respectively. It indicates that FreeSeg can adapt to more generalized scenarios.
ii) 
We also report the result of the fully supervised baseline, denoted as ``Full Sup.", which is trained on both seen and unseen classes.
Remarkably, FreeSeg is only 0.7\% and 5.2\% worse than the fully supervised baseline ``Full Sup." in seen and unseen classes on COCO, respectively.
iii) To compare with competitive methods that are only trained on VOC benchmark, we also report the result of FreeSeg in the same setting as previous work. The experimental results show that FreeSeg obtains 91.8\%/82.6\% mIoU on the seen and unseen classes, which outperforms ZSSeg by 12.6\%/4.5\%. It further proves that FreeSeg is both robust and excellent for handling multi-tasks and single task.

\begin{table}[tp]
    \centering
    \footnotesize
      \vspace{-10pt}
    \caption{Comparison with state-of-the-art methods in open vocabulary instance segmentation. mAP$^s$ and mAP$^u$ denote the mAP(\%) results of seen classes and unseen classes.}
      \vspace{-10pt}
    \begin{threeparttable}
    \resizebox{0.46\textwidth}{!}{
      \begin{tabular}{l|cc|ccc}
        \toprule
        \textbf{Method} & \textbf{mAP$^s$} & \textbf{mAP$^u$} & \textbf{mAP} & \textbf{AP$^{50}$} & \textbf{AP$^{75}$}   \\ 
        \midrule
        \multicolumn{6}{c}{\textit{COCO}} \\
        \midrule
          Full Sup. & 24.9 & 25.1 & 24.9 & 37.8 & 25.8  \\ \hline
        CLIP\cite{radford2021learning} & 8.5 & 2.6 & 7.9 & 11.8 & 7.5  \\
        ZSSeg\cite{xu2021simple} & 13.7 & 3.0 & 12.8 & 20.9 & 13.3  \\
        PL\cite{rahman2020improved} &  34.0 & 12.4 & - & - & - \\
        BLC\cite{zheng2020background} &  36.0 & 13.1 & - & - & - \\
        ZSI\cite{zheng2021zero} &  \textbf{38.7} &  13.6 & - & - & - \\
        \rowcolor{gray!10} Ours & 23.7 & \textbf{20.6} & \textbf{22.8} & \textbf{36.0} & \textbf{24.1} \\
        \midrule
        \multicolumn{6}{c}{\textit{ADE20K}} \\
        \midrule
        Full Sup. & 20.3 & 18.1 & 20.1 & 31.4 & 21.1  \\  \hline
        CLIP\cite{radford2021learning} & 5.6 & 3.5 & 5.4 & 7.7 & 5.8  \\
        ZSSeg\cite{xu2021simple} & 11.0 &  1.8 & 9.8 & 17.8 & 9.5  \\
        \rowcolor{gray!10} Ours & \textbf{16.3} & \textbf{15.4} & \textbf{16.2} & \textbf{25.3} & \textbf{16.9} \\
        \bottomrule
        \end{tabular}}
        \end{threeparttable}
        \label{tab:insseg}
        \vspace{-10pt}
    \end{table}  

\begin{table}[tp]
    \centering
    \footnotesize
      \vspace{-10pt}
    \caption{Comparison with state-of-the-art methods in PQ(\%), SQ(\%), RQ(\%) on open vocabulary panoptic segmentation. }
      \vspace{-10pt}
    \begin{threeparttable}
    \resizebox{0.48\textwidth}{!}{
      \begin{tabular}{l|ccc|ccc}
        \toprule
        \multirow{2}{*}{\textbf{Method}} &\multicolumn{3}{c|}{Seen} & \multicolumn{3}{c}{Unseen}\\ 
        & \textbf{PQ} & \textbf{SQ} & \textbf{RQ} & \textbf{PQ} & \textbf{SQ} & \textbf{RQ}  \\ 
        \midrule
        \multicolumn{7}{c}{\textit{COCO}} \\
        \midrule
        Full Sup. & 33.1 & 78.5 & 39.5 & 34.1 & 80.7 & 41.5 \\ \hline

        CLIP\cite{radford2021learning} & 14.3 & 71.5 & 18.4 & 9.2 & 70.3 & 11.6 \\
        ZSSeg\cite{xu2021simple} & 27.2 & 76.1 & 34.7 & 9.7 & 71.7 & 12.2 \\
        \rowcolor{gray!10} Ours & \textbf{31.4} & \textbf{78.3} & \textbf{38.9} & \textbf{29.8} & \textbf{79.2} & \textbf{37.6}
        \\
        \midrule
        \multicolumn{7}{c}{\textit{ADE20K}} \\
        \midrule
        Full Sup. & 35.3 & 78.1 & 43.9 & 30.6 & 74.2 & 37.5\\ \hline
        CLIP\cite{radford2021learning} & 9.5 & 62.8 & 12.1 & 3.4 & 61.1 & 4.7 \\
        ZSSeg\cite{xu2021simple} & 24.7 & 70.7 & 32.2 & 2.5 & 65.1 & 6.3 \\
        \rowcolor{gray!10} Ours & \textbf{32.8} & \textbf{78.2} & \textbf{40.4} & \textbf{25.4} & \textbf{75.2} & \textbf{30.6} \\
        \bottomrule
        \end{tabular}}
        \end{threeparttable}
        \label{tab:panseg}
        \vspace{-10pt}
    \end{table}

\paragraph{Open Vocabulary Instance Segmentation}
As shown in Tab.\ref{tab:insseg},  we compare the open vocabulary instance segmentation performance on COCO and ADE20K datasets, including ZSSeg\cite{xu2021simple}, PL\cite{rahman2020improved}, BLC\cite{zheng2020background}, and ZSI\cite{zheng2021zero}.
Since ZSSeg did not report the result on this task, we reproduce the results by training on the instance segmentation labels with the official code.
The variant ``CLIP" denotes the direct matching results with the pre-trained CLIP\cite{radford2021learning} text and visual encoder.
FreeSeg achieves 20.6\% mAP of unseen classes on COCO, which outperforms the best-performance method ZSI by +7.0\%  mAP. 
However, the mAP of the seen classes of ZSI is higher than FreeSeg.
It is because ZSI\cite{zheng2021zero} uses box-level supervision, which is more favorable for instance segmentation, while FreeSeg uses more general mask supervision for various segmentation tasks. 
In addition to COCO, FreeSeg also achieves promising results on ADE20k. For example, FreeSeg achieves 16.3\% / 15.4\%  mAP on seen / unseen classes, which outperforms the baseline CLIP by +10.7\%  and +11.9\% mAP.

\paragraph{Open Vocabulary Panoptic Segmentation}
Since few works study open vocabulary panoptic segmentation, we report the results of FreeSeg and the CLIP\cite{radford2021learning} baseline in Tab. \ref{tab:panseg}. We also re-implement ZSSeg\cite{xu2021simple} on panoptic segmentation labels to accomplish this task.
We observe that FreeSeg achieves 29.8\% PQ, 79.2\% SQ, and 37.6\% RQ of the unseen classes, outperforming ZSSeg by 20.1\%, 7.5\%, and 25.4\%, respectively. 
The main performance improvement comes from the unseen classes, indicating that this semantic segmentation-oriented method like ZSSeg is hard to generalize to other tasks, while our FreeSeg has noticeable generalization capability.
On the ADE20K dataset, FreeSeg also achieves the best results with 25.4\%, 75.2\%, and 30.6\% of unseen classes on PQ, SQ, and RQ, respectively. 
These above multi-task results prove the generalization ability of FreeSeg for unified open vocabulary segmentation tasks. 

\begin{table*}[ht]
    \caption{Ablation studies of the proposed modules on COCO datasets.}
    \vspace{-10pt}
    \label{tab: ablation}
    \resizebox{\textwidth}{!}{%
    \begin{tabular}{cc|c|c|cl|cl|ccclll}
    \toprule
      \multicolumn{2}{c|}{\textbf{Adaptive Prompt}} &
      \multirow{2}{*}{\textbf{\makecell{Context \\ Interaction}}} &
      \multirow{2}{*}{\textbf{\makecell{Prompt \\ Tuning}}} &
      \multicolumn{2}{c|}{\textbf{Semantic}} &
      \multicolumn{2}{c|}{\textbf{Instance}} &
      \multicolumn{6}{c}{\textbf{Panopnic}} \\ \cline{1-2} \cline{5-14} 
      \textbf{Class} &
      \textbf{Task} &
       &
       &
      \textbf{mIoU$^s$} &
      \textbf{mIoU$^u$} &
      \textbf{mAP$^s$} &
      \textbf{mAP$^u$} &
      \textbf{PQ$^s$} &
      \textbf{SQ$^s$} &
      \textbf{RQ$^s$} &
      \multicolumn{1}{c}{\textbf{PQ$^u$}} &
      \multicolumn{1}{c}{\textbf{SQ$^u$}} &
      \multicolumn{1}{c}{\textbf{RQ$^u$}} \\ 
      \midrule
     \XSolidBrush & \XSolidBrush & \XSolidBrush & \XSolidBrush & 38.4 & 4.9 & 19.2 & 0.7 & 25.9 & 70.4 & 32.0 &  0.1 & 0.2 & 0.1 \\
     \Checkmark & \XSolidBrush & \XSolidBrush  & \XSolidBrush & 39.0 & 38.5 & 20.1 & 8.8 & 26.2 & 70.7 & 32.5 & 12.0 & 62.6 & 15.3 \\
     \Checkmark & \Checkmark &\XSolidBrush  &\XSolidBrush  & 40.8 & 41.3 & 22.2 & 11.8 & 28.5 & 74.3 & 35.8 & 15.7 & 67.4 & 19.8 \\
     \Checkmark & \Checkmark & \Checkmark & \XSolidBrush  & 42.1 & 42.6 & 23.7 & 13.9 & 30.0 & 76.5 & 37.3 & 18.1 & 70.5 & 23.2 \\
     \Checkmark & \Checkmark & \Checkmark & \Checkmark & \textbf{41.9} & \textbf{43.3} & \textbf{23.9} & \textbf{14.6} & \textbf{30.4} & \textbf{76.7} & \textbf{38.1} & \textbf{19.2} & \textbf{71.4} & \textbf{24.1}  \\ 
    \bottomrule
    \end{tabular}%
    }
    \end{table*}

\begin{table}[tp]
    \centering
    \footnotesize
      \vspace{-5pt}
    \caption{Generalization performance (in\%) of the open vocabulary segmentation on cross datasets.}
      \vspace{-5pt}
    \begin{threeparttable}
    \resizebox{0.48\textwidth}{!}{
      \begin{tabular}{lc|c|ccc}
        \toprule
        \textbf{Method} & \textbf{mIoU} & \textbf{mAP} & \textbf{PQ} & \textbf{SQ} & \textbf{RQ}  \\ 
        \midrule
        \multicolumn{6}{c}{\textit{COCO $\rightarrow$ ADE20K}} \\
        \midrule
        CLIP\cite{radford2021learning} & 13.8 & 3.9 & 8.2 & 53.1 & 10.5 \\
        Lseg+~\cite{ghiasi2022scaling} & 13.0 & - & - & - & - \\
        OpenSeg~\cite{ghiasi2022scaling} & 15.3 & - & - & - & - \\
        ZSSeg\cite{xu2021simple} & 16.4 & 4.0 & 9.3 & 58.0 & 12.2\\
        MaskCLIP\cite{ding2022open} & 23.7 & 5.9 & 15.1 & 70.4 & 19.2 \\
        \rowcolor{gray!10} Ours & \textbf{24.6} & \textbf{6.5} & \textbf{16.3} & \textbf{71.8} & \textbf{21.6} \\
        \midrule
        \multicolumn{6}{c}{\textit{ADE20K $\rightarrow$ COCO}} \\
        \midrule
        CLIP\cite{radford2021learning} & 14.7 & 2.7 & 8.1 & 66.3 & 11.0\\
        ZSSeg\cite{xu2021simple} & 17.7 & 4.3 & 11.2 & 66.5 & 14.9\\
        \rowcolor{gray!10} Ours & \textbf{21.7} & \textbf{6.6} & \textbf{16.5} & \textbf{72.0} & \textbf{21.6} \\
        \bottomrule
        \end{tabular}}
        \end{threeparttable}
        \label{tab:generalization}
    \end{table}  

\begin{table}[tp]
    \centering
    \small
      \vspace{-8pt}
    \caption{Generalization performance (in\%)  of the open vocabulary semantic segmentation on VOC2012 datasets.}
      \vspace{-8pt}
    \begin{threeparttable}
    \resizebox{0.48\textwidth}{!}{
        \begin{tabular}{lcc}
        \toprule
        \textbf{Method} & \multicolumn{2}{c}{\textbf{mIoU}}  \\ 
        \midrule
        & \textit{COCO $\rightarrow$ VOC2012} & \textit{ADE20K $\rightarrow$ VOC2012} \\
        \midrule
        CLIP\cite{radford2021learning} & 71.6  & 67.1   \\
        ZSSeg\cite{xu2021simple} & 82.1   & 69.2  \\
        \rowcolor{gray!10} Ours & \textbf{91.9} & \textbf{80.1}  \\
        \bottomrule
        \end{tabular}}
        \end{threeparttable}
        \label{tab: generalization_voc}
    \end{table}

\subsection{Generalization Analysis}
We evaluate the generalization of FreeSeg across datasets. Namely, we train FreeSeg on COCO ($G_{coco}$) or ADE20K ($G_{ade}$) and directly test it on other datasets without finetuning.
This cross-dataset evaluation is non-trivial because of the significant differences in the data distribution and domains. 
In this setting, all classes of the target dataset are regarded as unseen classes, and we report the segmentation performance of all unseen classes for three segmentation tasks.
As shown in Tab. \ref{tab:generalization}, FreeSeg achieves 24.6\% mIoU semantic segmentation, 6.5\% mAP instance segmentation, 16.3\% PQ, 71.8\% SQ, 21.6\% RQ panoptic segmentation results on ADE20K with $G_{coco}$, which outperforms the SOTA method MaskCLIP\cite{ding2022open} with 0.9\% mIoU, 0.6\% mAP, 1.2\% PQ, and 2.4\% RQ, respectively. 
FreeSeg also obtains the best performance when validating $G_{ade}$ on COCO  datasets, achieving 21.7\% mIoU, 6.6\% mAP, 16.5\% PQ, 72.0\% SQ, and 21.6\% RQ for semantic, instance, and panoptic segmentation, respectively. 
The generalization results on VOC2012  with $G_{coco}$  and $G_{ade}$ also verify the transferability of FreeSeg in Tab.\ref{tab: generalization_voc}

\subsection{Ablation Study}

\paragraph{Component Analysis}

We conduct ablation studies to analyze essential components of FreeSeg on COCO datasets in Tab.\ref{tab: ablation}. Note that the self-training technique is not applied in these ablations.
The primary vision model achieves an inferior performance of 4.9\% mIoU and 0.7\% mAP on the unseen classes without any text guidance. 
By introducing the adaptive class prompt, the performance is improved significantly on COCO, especially for the unseen classes.
Then the adaptive task prompt and semantic context interaction module is gradually inserted into the framework, which brings out the performance improvement of 2.8\% and 1.3\% mIoU on COCO dataset, respectively. 
Furthermore, the experimental results show that test time prompt tuning also improves the unseen classes' performance during inference.

We also explore the effectiveness of the proposed modules on the open vocabulary instance and panoptic segmentation and obtain a highly consistent conclusion with semantic segmentation.
It demonstrates that adaptive prompt learning promotes FreeSeg to capture task-aware and category-sensitive characteristics. 
The semantic context interaction and test-time prompt-tuning help to improve the cross-modal alignment of visual and text features.

\begin{figure*}[tp]
    \centering
    \begin{center}
        \includegraphics[width=0.98\textwidth]{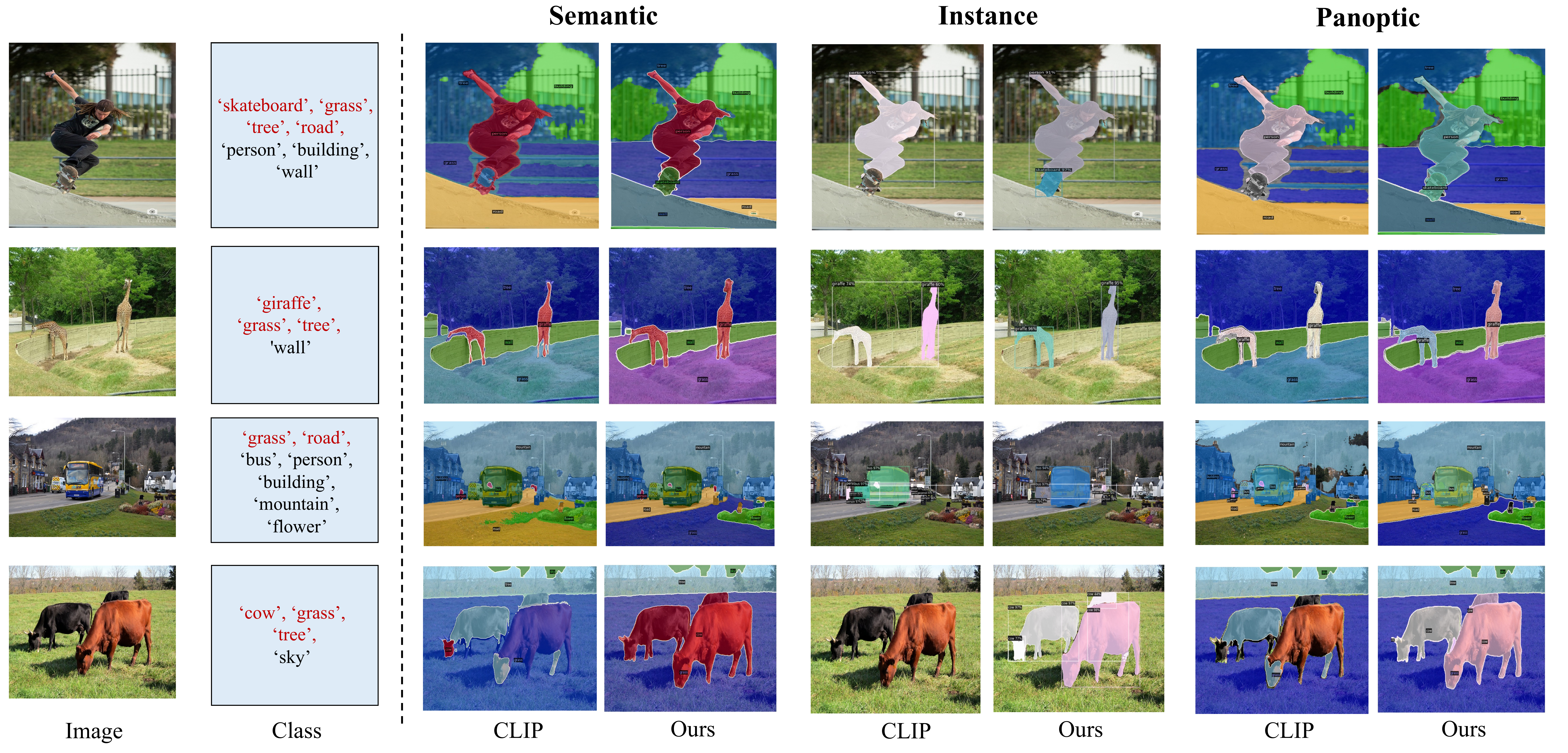}
    \end{center}
        \vspace{-20pt}
    \caption{Qualitative results of the multi-task open vocabulary segmentation. We compare the segmentation results of the proposed FreeSeg and CLIP\cite{radford2021learning}. The class column represents the class names, where \textcolor{red}{red} and black words denote the \textcolor{red}{unseen} and seen classes, respectively. }
    \label{fig: visualization}
    \vspace{-5pt}
    \end{figure*}

\begin{table*}[tp]
    \centering
    \footnotesize
    \caption{Comparison of different training paradigms and prompt solutions on COCO.
    }
    \vspace{-8pt}
    \begin{threeparttable}
    \resizebox{0.98\textwidth}{!}{
        \begin{tabular}{cc|cc|cc|ccccccc}
        \toprule
        \multicolumn{2}{c|}{\multirow{2}{*}{\textbf{Method}}} & \multicolumn{2}{c|}{\textbf{Semantic}} & \multicolumn{2}{c|}{\textbf{Instance}} & \multicolumn{6}{c}{\textbf{Panoptic}} \\
        \cline{3-12}
        & & \textbf{mIoU$^s$} & \textbf{mIoU$^u$} & \textbf{mAP$^s$} & \textbf{mAP$^u$} & \textbf{PQ$^s$} & \textbf{SQ$^s$} & \textbf{RQ$^s$} & \textbf{PQ$^u$} & \textbf{SQ$^u$} & \textbf{RQ$^u$} \\ 
        \midrule
        \multirow{2}{*}{\textbf{Train Paradigm}} & \multicolumn{1}{|c|}{Single-Task} & 41.3 & 42.9 & 24.1 & 12.7 & 30.1 & 75.0 & 37.6 & 17.5 & 69.7 & 21.1 \\
        & \multicolumn{1}{|c|}{Multi-Task} & 41.9 & 43.3 & 23.9 & 14.6 & 30.4 & 76.7 & 38.1 & 19.2 & 71.4 & 24.1 \\
        \midrule
        \multirow{2}{*}{\textbf{Prompt}} & \multicolumn{1}{|c|}{Fixed} & 38.5 & 33.7 & 21.4 & 8.2 & 26.1 & 73.0 & 32.4 & 12.1 & 63.6 & 17.5 \\
        & \multicolumn{1}{|c|}{Adaptive} & 41.9 & 43.3 & 23.9 & 14.6 & 30.4 & 76.7 & 38.1 & 19.2 & 71.4 & 24.1 \\
        \bottomrule 
        \end{tabular}}
        \end{threeparttable}
        \label{tab: multitask}
    \end{table*}

\paragraph{Multi-Task Analysis.}
To validate the advantages of multi-task learning in FreeSeg, we compare the results of the unified multi-task training with the single-task training for specific tasks. 
As shown in Tab.\ref{tab: multitask}, all the results from the multi-task row are obtained from one unified model, while the single-task results are from three individual models. All results are obtained without the self-training technique. 
Multi-task training achieves 41.9\% and 43.3\%  mIoU for the seen and unseen classes on open vocabulary semantic segmentation, suppressing the performance of the single-task model. 
Open-vocabulary instance and panoptic segmentation also show consistent results as semantic segmentation, especially in the performance of unseen classes. 
FreeSeg improves all metrics of unseen classes on all tasks, proving that the multi-task training scheme can efficiently improve the generalization of the networks.
Furthermore, the unified open vocabulary model conducts a one-shot training procedure with multi-task labels, which achieves superior performance while reducing nearly 2/3 of training costs.

\paragraph{Adaptive Prompt Analysis.}
We compare the results of different prompt settings to verify the importance of the adaptive prompt for open vocabulary segmentation in Tab.\ref{tab: multitask}. 
The fixed template prompt uses the template sentence ``A photo of \{\textit{class}\}." where \{\textit{class}\} is placed in specific class names. The task name is filled into the template ``for \{\textit{task}\}." to get the fixed task prompt. Then the task prompt and the class prompt are encoded into the text features.
As shown in Tab.\ref{tab: multitask}, the adaptive prompt brings out 3.4\% and 9.6\% mIoU performance improvement than the fixed prompt regarding seen and unseen classes, respectively. 
Similarly, the adaptive prompt outperforms the fixed prompt by 0.9\% and 2.2\% mAP on instance segmentation and by 8.2\% and 7.5\% PQ on panoptic segmentation. 
It reveals that adaptive prompt facilitates the prompt to capture task-aware and category-sensitive concepts via learnable parameters.

\subsection{Qualitative results}
We visualize the qualitative results of the unified open vocabulary segmentation in Fig.\ref{fig: visualization}. It can be observed that CLIP fails to segment the instances of some unseen classes like ``cow'' and ``skateboard'' in the first and fourth images. However, FreeSeg accurately segments the unseen class regions such as ``giraffe'' or ``grass'' for semantic segmentation.
These figures show our capability of specifying arbitrary classes in instance and panoptic segmentation.
These results demonstrate that FreeSeg is capable of generalizing to arbitrary segmentation categories in universal segmentation tasks.

\section{Conclusion}
In this paper, we provide a universal framework, \ie, FreeSeg to accomplish unified open-vocabulary segmentation.
To the best of our knowledge, we offer the first attempt to employ a single model with the same architecture and inference parameters to accomplish open-vocabulary semantic, instance, and panoptic segmentation.
Compared with single-task training, FreeSeg successfully reduced the training cost by about two-thirds and achieved better generalization performance.  Only one unified model is needed in real-scene deployment, reducing the inference procedure's computational capacity, memory cost, and bandwidth. 
We believe our work can provide inspired insight and suggest a new path forward in open-vocabulary segmentation.

{\small
\bibliographystyle{ieee_fullname}
\bibliography{egbib}
}

\end{document}